\documentclass[10pt,twocolumn,letterpaper]{article}
\usepackage{3dv}
\usepackage{times}
\usepackage{epsfig}
\usepackage{graphicx}
\usepackage{amsmath}
\usepackage{amssymb}
\usepackage{bm}
\usepackage{enumerate}
\usepackage{color}
\usepackage{multirow}
\usepackage{array}
\usepackage{bigstrut}
\usepackage{algorithmic}
\usepackage{algorithm}
\usepackage{arydshln}

\usepackage[caption=false,font=footnotesize]{subfig}
\graphicspath{{figures/}}

\newcommand{\comment}[1]{}
\newcommand{\deflen}[2]{%
    \expandafter\newlength\csname #1\endcsname
    \expandafter\setlength\csname #1\endcsname{#2}%
}

\def\threedvPaperID{153} 

\ifthreedvfinal\fi
\begin{document}

\title{DSR: Direct Self-rectification for Uncalibrated Dual-lens Cameras}

\author{Ruichao Xiao\thanks{R. Xiao and W. Sun contributed equally to this work.} \and Wenxiu Sun\footnotemark[1] \and Jiahao Pang \and Qiong Yan \and Jimmy Ren \and
SenseTime Research\\
{\tt\small \{xiaoruichao, sunwenxiu, pangjiahao, yanqiong, rensijie\}@sensetime.com}
}
\maketitle
\thispagestyle{empty}

\begin{abstract}
With the developments of dual-lens camera modules, depth information representing the third dimension of the captured scenes becomes available for smartphones.
It is estimated by stereo matching algorithms, taking as input the two views captured by dual-lens cameras at slightly different viewpoints. Depth-of-field rendering (also be referred to as synthetic defocus or bokeh) is one of the trending depth-based applications.
However, to achieve fast depth estimation on smartphones, the stereo pairs need to be rectified in the first place.
In this paper, we propose a cost-effective solution to perform stereo rectification for dual-lens cameras called direct self-rectification, short for DSR\footnote{Code is avaliable at \href{https://github.com/garroud/self-rectification}{github.com/garroud/self-rectification}}
.
It removes the need of individual offline calibration for every pair of dual-lens cameras.
In addition, the proposed solution is robust to the slight movements, {\it e.g.}, due to collisions, of the dual-lens cameras after fabrication.
Different with existing self-rectification approaches, our approach computes the homography in a novel way with zero geometric distortions introduced to the master image.
It is achieved by directly minimizing the vertical displacements of corresponding points between the original master image and the transformed slave image.
Our method is evaluated on both realistic and synthetic stereo image pairs, and produces superior results compared to the calibrated rectification or other self-rectification approaches.

\end{abstract}


\section{Introduction}
\label{sec:intro}
Image rectification is a crucial component for fast stereo matching, especially for smartphones or other platforms with limited computational resources.
Taking rectified images as inputs, the correspondence matching becomes restricted in the same scan-line, which largely reduces the computational demand.

\deflen{figresslf}{148pt}
\begin{figure*}[!t]
        \centering
        \subfloat[(a) Master Image]
        {\includegraphics[height=\figresslf]{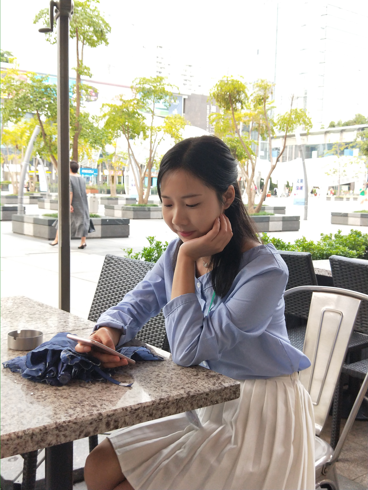}}\hspace{1pt}
        \subfloat[(b) Direct Self-rectification (details)]
        {\includegraphics[height=\figresslf]{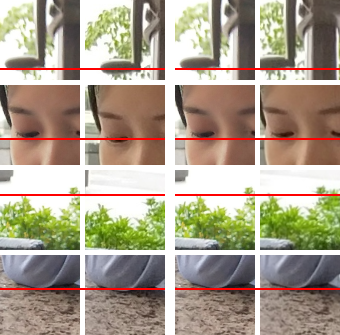}}\hspace{1pt}
        \subfloat[(c) Stereo Matching]
        {\includegraphics[height=\figresslf]{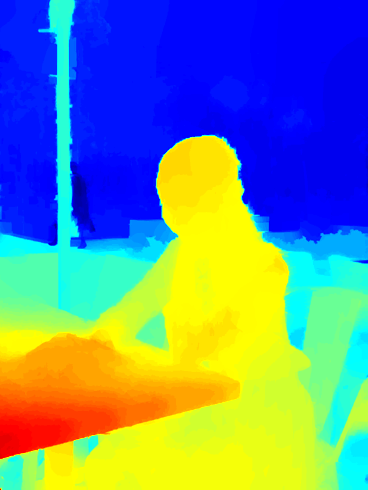}}\hspace{1pt}
        \subfloat[(d) Depth-of-field Rendering]
        {\includegraphics[height=\figresslf]{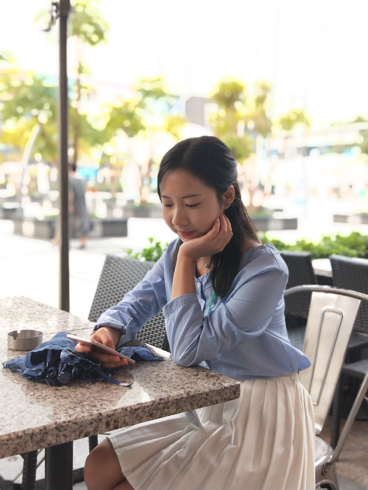}}
    \caption{Stereo rectification is important and its application on stereo matching and depth-of-field rendering.
    }
    \label{fig:example}
\end{figure*}

Traditionally, image rectification is the process of applying a pair of projective transformations ({\it i.e.,} homographies) to a pair of images ({\it i.e.,} the master image and the slave image) whose epipolar geometry is known, so that the epipolar lines in the original images map to horizontally aligned lines in the transformed images.
Examples of co-located image fragments before and after applying rectification are shown in Fig.~\ref{fig:example}(b).
However, existing image rectification paradigm has two shortcomings.
First, to get the epipolar geometry, offline calibration or online calibration is required. Though capturing the stereo images using dual-lens cameras is straight-forward, calibrating them offline is cumbersome, in terms of: 1) setting up the calibration environment; 2) individually calibration of each dual-lens cameras; and 3) fix the dual-lens module after calibration, otherwise the calibrated parameters would have degraded usage or even not useful at all.
Second, applying a pair of projective transformations brings two side-effects which are not desirable for stereo vision tasks. One is that there are undefined regions in the transformed image which may cause matching ambiguities during stereo matching. The other is that there is geometric distortion in the transformed image, which are not preferred for high-quality depth-based applications, such as depth-of-field rendering (Fig.\,1(d)).

Notice that it is a commonly adopted set-up that dual-lens cameras are \emph{laterally displaced}, such as those equipped on smartphones or on robots. Without limiting to the physical dual-lens cameras, if a hand-held camera moves horizontally and capture the scenes at two time instants, we also interpret this setting as dual-lens cameras in a general case.
Based on this configuration, we propose a novel self-rectification algorithm for uncalibrated stereo images called \emph{DSR}.
Our proposal keeps the master image unchanged and applies homography only on the slave image.
Moreover, no additional information is required except for the stereo images.
To achieve these features, we carefully examine the feasibility in the laterally displaced stereo and formulate the self-rectification as a regression problem without explicitly knowing the epipolar geometry.
Particularly, the vertical displacement error of detected correspondence pairs between the master image and the slave image is first minimized.
Then a shearing transformation is computed to minimize the amount of geometric distortion of the transformed slave image.
Lastly, the transformed slave image is shifted horizontally to make the largest disparity being $0$ to facilitate the follow-up stereo matching algorithm.

To demonstrate the superiority of the proposed self-rectification algorithm, we evaluate the proposed DualRec on synthetic stereo images with similar settings of smart-phones, while each synthesized pair has slightly different configurations in terms of camera parameters. We also evaluate on realistic stereo images acquired from dual-lens smartphones captured in various scenarios.
Our method is applicable to vertically displaced cameras as well by switching $x$- and $y$-axis during problem formulation.
Experimentation on either synthetic stereo images or on realistic stereo images shows that, our approach provides promising results, out-performing prior state-of-the-art solutions both quantitatively and qualitatively.

Differing with previous approaches that targeted for rectification of uncalibrated stereo or calibrated stereo, our contributions are:
\begin{itemize}
    \item We find that for laterally displaced stereo cameras, the homographies can be computed in a novel way without requiring the epipolar geometry, thus reducing the calibration cost for every pair of the dual-lens cameras.
    \item We present a self-rectification approach introducing zero geometric distortion to the master image, which brings stable results for stereo matching and for depth-based image applications.
    \item We have carefully examined the usage and limitation of the proposed method, showing that it is applicable for a wide range of dual-lens cameras which are laterally displaced.
\end{itemize}
Our paper is organized as follows. We review related works in Section\,\ref{sec:related}. In Section\,\ref{sec:method}, we elaborate our proposed self-rectification. The experimental results and conclusions are presented in Section\,\ref{sec:results} and Section\,\ref{sec:conclude}, respectively.

\section{Related Works}
\label{sec:related}
Rectification is a classical problem in stereo vision. 
If the intrinsic and extrinsic camera parameters are pre-calibrated, one may adopt a compact algorithm, proposed by Fusiello~{\it et. al}~\cite{fusiello00}, to find the homographies in a few lines of code. 
However, such camera parameters are often not available and are volatile due to any mechanical misalignment of the stereo rig.

To resolve this headache, finding the epipolar geometry ({\it i.e.,} fundamental matrix or essential matrix) which reflects all geometric information contained in the two images is an alternative to a full calibration. 
An overview of the relevant techniques can be found in \cite{zhang98}. 
Given such epipolar geometry, Loop~{\it et al.}~\cite{loop99} estimated homographies by decomposing them into a specialized projective transform, a similarity transform and a shearing transform, so as to reduce the geometric distortion of the rectified image pairs.
Gluckman and Nayar~\cite{gluckman01} proposed a rectification method to minimize the re-sampling effect, namely the loss of pixels due to under-sampling and the creation of new pixels due to over-sampling. 
Isgro and Trucco~\cite{isgro99} proposed an approach without explicit computation of the fundamental matrix by rectifying homographies directly from image correspondences. Fusiello and Irsara~\cite{fusiello2011quasi} proposed a new Quasi-Euclidean rectification algorithm by adopting Sampson error~\cite{hartley2003multiple} to better capture geometric reprojection error.
More recently, Zilly~{\it et al.}\cite{zilly10} proposed a technique to jointly estimate the epipolar geometry and the rectification parameters for almost parallel stereo rigs. 
Although quite a few methods are proposed to reduce the unwanted geometric distortions, the homographies can be easily affected by the uncertainties in the epipolar geometry, which may result in unstable or unreasonable disparities after stereo matching.

As opposed to previous approaches that require calibrated stereo rig or estimated epipolar geometry, we propose a practical solution called DSR for the laterally displaced stereo 
where the stereo cameras have barley horizontal displacement. 
By estimating a homography only for the slave image with minimized geometric distortion, we achieve state-of-the-art rectification results with superior accuracy. It is also demonstrated very effective for the follow-up stereo matching algorithms.

\section{Methodology}
\label{sec:method}
In this section, we elaborate the proposed self-rectification method in detail. As a convention, a matrix and a vector will be denoted respectively by boldface uppercase letter ({\it e.g.,} $\bf{H}$) and lowercase letter ({\it e.g.,} $\bf{p}$), and a scalar will be denoted
by an italic upper or lowercase letter ({\it e.g.,} $N$ or $n$). For ease of representation, we adopt homogeneous coordinate system as commonly used in 3D computer vision, where image points in 2D are represented by 3D column vectors, {\it e.g.,} ${\bf{p}} = [x\ y\ 1]^{\rm T}$.  As in homogeneous coordinate, points are scale-invariant, hence $[x\ y\ 1]^{\rm T}$ and $[\alpha x\ \alpha y\ \alpha]^{\rm T}$ denote the same point. We will firstly describe the dual-lens cameras and propose our small-drift assumption in Section \ref{subsec:lds}. Based on this assumption, the self-rectification approach called DSR is then presented in Section \ref{subsec:alg}. Lastly, in Section \ref{subsec:small-drift}, we analyze from simulated experiments to further quantify the small-drift assumption. 

\subsection{Motivations and Assumptions}
\label{subsec:lds}

\begin{figure}[t!]
\centering
    \subfloat[(a) General stereo \label{fig:geo_general}]{\includegraphics[width=0.5\columnwidth]{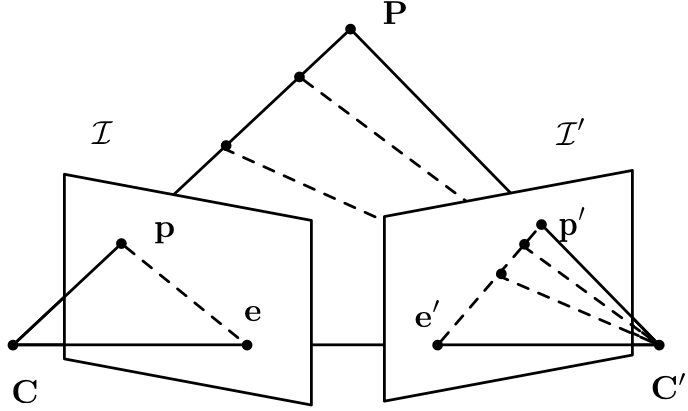}}\hspace{20pt}
    \subfloat[(b) Laterally displaced stereo \label{fig:geo_lateral}]{\includegraphics[width=0.7\columnwidth]{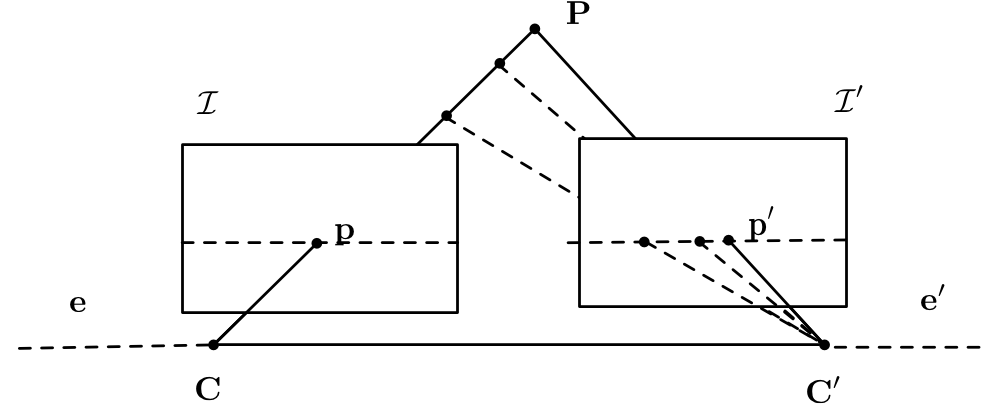}}\hspace{1pt}
\caption{Epipolar geometry between a pair of images.}
\label{fig:geo_generals}
\end{figure}

\comment{
}

Dual-lens cameras exist widely, for example, those mounted on robots or mobile phones. Without limiting to the physical stereo cameras, if a hand-held camera moves horizontally and capture the scenes at two time instants,  we also interpret this setting as stereo cameras in a general sense. In this system, the line connecting two camera centers is almost parallel to the image planes and the scanlines ({\it i.e.,} $x$-axis) in the image plane. In Fig.~\ref{fig:geo_general} and Fig.~\ref{fig:geo_lateral}, we plot the epipolar geometry for a general stereo vision system and for a laterally displaced stereo vision system, respectively, where $\mathcal{I}$ is the master image plane, $\mathcal{I'}$ is the slave image plane, $\bf{C}$ and $\bf{C'}$ are camera centers,  $\bf{P}$ is a point in the 3D space with $\bf{p}$ and $\bf{p'}$ being its projections on the image planes. The line $\mathbf{CC'}$ intersects the image planes at locations $\mathbf{e}$ and $\mathbf{e'}$ which are termed as epipoles. If $\mathbf{P}$ varies its location along the line ${\bf Cp}$, its projection on the slave image shall lies on the epipolar line $\mathbf{e'p'}$, and vice versa. Thus, $\mathbf{ep}$ and $\mathbf{e'p'}$ forms a pair of corresponding lines.
The objective of rectification is to push the epipoles, $\mathbf{e}$ and $\mathbf{e}'$, to infinity, such that the corresponding lines in the image planes would become the same scanline.

In laterally displaced stereo vision system, Fig.~\ref{fig:geo_lateral}, $\bf{CC'}$ is almost parallel to the master image plane $\mathcal{I}$, making the epipole $\bf{e}$ far from the image center. In the perfect case, where $\mathbf{CC'}$ is strictly parallel to the master image plane, the master image does not need any projective transformations as its epipolar lines are already parallel to the scanline. 
However, in practice, the stereo rig may suffer from small perturbations, leading to relative rotation and/or relative translation with regard to the perfect case. We analyze these perturbations in the following. For simplification, we use the perfect case as a reference, and denote the perfect slave image center and its image plane as $\mathbf{C'}$ and $\mathcal{I'}$, respectively. The imperfect slave image center and its image plane are denoted as $\mathbf{C''}$ and $\mathcal{I''}$, respectively.

{\bf A. Relative Rotation.} In this case, the camera plane $\mathcal{I''}$ has only relative rotation with respect to its perfect case $\mathcal{I'}$, as shown in Fig.~\ref{fig:geo_cocenter},  where $\mathbf{C'}$ and $\mathbf{C''}$ are co-located. It can be proved that there always exist a homography between $\mathcal{I''}$ and $\mathcal{I'}$. For an arbitrary point $\mathbf{P} = [X\ Y\ Z\  1]^{\rm T}$ in 3D space, the projected points $\mathbf{p'}$ and $\mathbf{p''}$ are
\begin{equation}
\mathbf{p'} \sim \mathbf{K'}[\mathbf{I}|\mathbf{0}][X\ Y\ Z\ 1]^{\rm T} = \mathbf{K'}[X\ Y\ Z]^{\rm T},
\label{eq:proj_perfect}
\end{equation}
\begin{equation}
\mathbf{p''} \sim \mathbf{K''}[\mathbf{R''}|\mathbf{0}][X\ Y\ Z\ 1]^{\rm T} = \mathbf{K''}\mathbf{R''}[X\ Y\ Z]^{\rm T},
\label{eq:proj_rot}
\end{equation}
where $\mathbf{K'}$ and $\mathbf{K''}$ are the intrinsic camera matrices, $\mathbf{I}$ is an identity matrix, $\mathbf{0}$ is a vector of zeros indicating no translation, $\mathbf{R''}$ is the relative rotation matrix. Substituting (\ref{eq:proj_perfect}) into (\ref{eq:proj_rot}), we can derive
\begin{equation}
\mathbf{p''} \sim \mathbf{K''}\mathbf{R''}(\mathbf{K'})^{-1}\mathbf{p'}.
\label{eq:proj_homo}
\end{equation}
Therefore in this case, the homography can be computed without approximation.

\comment{
\begin{figure}[t]
\centering
    \includegraphics[width=0.6\columnwidth]{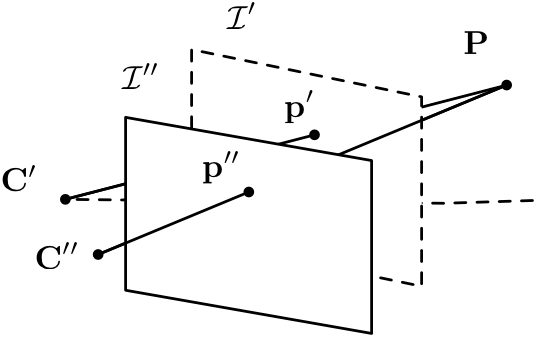}
\caption{A stereo rig with relative translation.}
\label{fig:geo_parallel}
\end{figure}
}

\begin{figure}[t!]
\centering
    \subfloat[(a) Relative rotation \label{fig:geo_cocenter}]{\includegraphics[width=0.64\columnwidth]{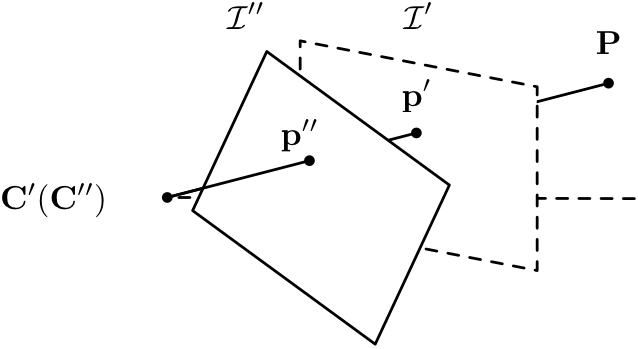}}\hspace{60pt}
    \subfloat[(b) Relative translation \label{fig:geo_parallel}]{\includegraphics[width=0.54\columnwidth]{geo_parallel.png}}\hspace{1pt}
\caption{The camera plane $\mathcal{I''}$ has relative rotation or relative translation with its perfect case $\mathcal{I'}$.}
\label{fig:geo_generals}
\end{figure}

\comment{
\begin{figure}[t]
\centering
    \includegraphics[width=0.7\columnwidth]{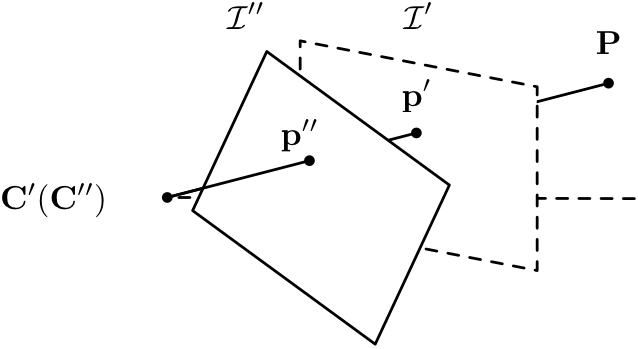}
\caption{A stereo rig with relative rotation.}
\label{fig:geo_cocenter}
\end{figure}
}

{\bf B. Relative Translation.} In this case, the image planes $\mathcal{I'}$ and $\mathcal{I''}$ are parallel, however, the camera center $\mathbf{C''}$ is off from its perfect counterpart  $\mathbf{C'}$ by a tiny shift $\mathbf{t} = [0\,t_y\,t_z]^{\rm T}$, shown in Fig.~\ref{fig:geo_parallel}. We ignore the shift in $x$-axis, as we can always find a perfect reference with the same $x$-coordinate. Then the projection becomes
\begin{equation}
\mathbf{p''} \sim \mathbf{K''}[\mathbf{I}|\mathbf{t}][X\ Y\ Z\ 1]^{\rm T} = \mathbf{K''}[X\ Y\ Z]^{\rm T} + \mathbf{K''}\mathbf{t}.
\label{eq:proj_tsl}
\end{equation}
When $t_y \ll Y$ and $t_z \ll Z$ are satisfied, $\mathbf{K''}\mathbf{t}$ can be ignored. We call this assumption as \emph{small-drift} assumption. In this case, an approximated homography can be found between the two image planes. Note that as $\mathbf{C}$ and $\mathbf{C'}$ has no translation in the $y$- and $z$-axis in 3D, $t_y$ and $t_z$ are also the amount of relative translation between the master image and the slave image.

{\bf C. Relative Rotation and Relative Translation.} If both relative rotation and translation exists, one can first rotate one image plane to make them parallel without loss of generality. Then the problem reduces to the case of relative translation with the same approximation made.

\subsection{The DSR Algorithm} 
\label{subsec:alg}
As discussed in Section~\ref{subsec:lds}, if the small-drift assumption satisfies, we can find an approximated homography to align the master image and the slave image. 
This assumption ({\it i.e.,} $t_y \ll Y$ and $t_z \ll Z$) can be easily satisfied for dual-lens smartphones, since the cameras are fixed in phones with very small relative shifts.
As opposed to prior works that use epipolar geometry between a pair of images to find homographies for both images, in this paper, we only need to find one homography for the slave image, without altering the master image.

Given a set of corresponding points that are identified by a feature matching approach or by manual labor, we aim at finding the appropriate transformation matrix. Let $\{\mathbf{p}_i, \mathbf{p}_i'\}_{i=1}^{N}$ be the set of corresponding points, where $\mathbf{p}_i = [x_i\ y_i\ 1]^{\rm T}$ and $\mathbf{p'}_i = [x'_i\ y'_i\ 1]^{\rm T}$.
Let $\bf{H}$ be the transformation matrix for the slave image, where $h_{11}, h_{12}, ..., h_{33}$ are its 9 entries:
\[ \bf{H} = 
    \begin{bmatrix}
        h_{11} & h_{12} & h_{13} \\
        h_{21} & h_{22} & h_{23} \\
        h_{31} & h_{32} & h_{33}
    \end{bmatrix}.
\]
As in homogeneous coordinates, multiplying a non-zero scalar does not change the homography, we simply let $h_{33}=1$. 
To find the remaining 8 elements in $\mathbf{H}$, we decompose it into three matrices:
\begin{equation}
\mathbf{H} = \mathbf{H}_k \mathbf{H}_s \mathbf{H}_y,
\end{equation}
where $\mathbf{H}_y$ aligns the $y$-coordinate of corresponding pixels, $\mathbf{H}_s$ serves as a shearing matrix to reduce the geometric distortion of the transformed slave image, $\mathbf{H}_k$ is used to shift image horizontally to guarantee negative disparities for the intention of stereo matching. We present the computation for each of them as follows.

{\bf Computation of $\mathbf{H}_y$.} We compute $\mathbf{H}_y$ by minimizing the vertical alignment error between the master image and the transformed slave image. As the $y$-coordinate of the transformed points are determined by the last two rows of $\mathbf{H}_y$, we define it as 
\[ {\bf H}_y = 
    \begin{bmatrix}
        1 & 0 & 0\\
        h_{21} & h_{22} & h_{23} \\
        h_{31} & h_{32} & 1
    \end{bmatrix}.
\]
Let $\mathbf{h}_2$ and $\mathbf{h}_3$ be the second row and the third row of the matrix $\mathbf{H}_y$. Then we minimize the vertical alignment error by solving the following problem,
%
\begin{equation}
\mathbf{H}_y^{\star} =  \arg \min \sum_i{(\frac{{\bf h}_2 \cdot {\bf p'}_i}{{\bf h}_3 \cdot {\bf p'}_i} - y_i)^2}.
\label{eq:argmin_H_y}
\end{equation}

By changing (\ref{eq:argmin_H_y}) as
\begin{equation}
\mathbf{H}_y^{\star} = \arg \min \sum_i{({\bf h}_2 \cdot {\bf p'}_i - {\bf h}_3 \cdot {\bf p'}_i \cdot y_i)^2},
\label{eq:regress_H_y}
\end{equation}
it becomes a multi-variable regression problem. It follows that the regression function is:
\begin{equation}
\underbrace{
    \begin{bmatrix}
        x'_1 & y'_1 & 1 & - x'_1y_1 & -y'_1y_1   \\
        x'_2 & y'_2 & 1 & - x'_2y_2 & -y'_2y_2   \\
        \vdots  & \vdots & \vdots & \vdots& \vdots  \\
        x'_n & y'_n & 1 & - x'_ny_n & -y'_ny_n    
    \end{bmatrix}}_{\mathbf{A}}\bm\cdot
    \underbrace{\begin{bmatrix}
        h_{21} \\ h_{22} \\ h_{23} \\ h_{31} \\ h_{32} 
    \end{bmatrix}}_{\mathbf{h}}
    =
    \underbrace{\begin{bmatrix}
        y_1 \\ y_2 \\ \vdots \\ y_n 
    \end{bmatrix}}_{\mathbf{y}}.
    \label{eq:ax=b}
\end{equation} 
Hence the optimal solution is given by
\begin{equation}
\mathbf{h^{\star}} = \mathbf{A}^{\dagger}\mathbf{y},
\label{eq:hy}
\end{equation}
where ${\bf A}^{\dagger}$ denotes the pseudo inverse of ${\bf A}$.

Additionally, we apply RANSAC to make the proposed algorithm robust to possible outliers in the matched keypoints. In particular, as described in~Algorithm~\ref{alg:sRec}, we compute a temporary alignment matrix $\mathbf{\widehat{H}}_y$ on randomly selected $M$ pairs of matched points. Let $y_i$ be the y-coordinate of $\mathbf{p}_i$, and $\tilde{y}_i$ be the y-coordinate of the warped points $\mathbf{\widehat{H}}_y \mathbf{p}_i'$. Such points pair is counted as \emph{inlier} if the vertical displacement error $|y_i - \tilde{y}_i|$ is less than a threshold $\epsilon$. We iterate this process for at most $T$ times, and select the one with largest percentage of inliers among the computed $\mathbf{\widehat{H}}_y$ as $\mathbf{H}_y$.

\comment{
\begin{figure}[!t]
        \centering
        {\includegraphics[width=0.46\linewidth]{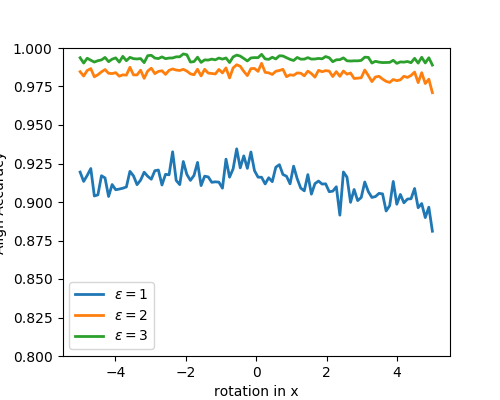}}\hspace{1pt}
        {\includegraphics[width=0.46\linewidth]{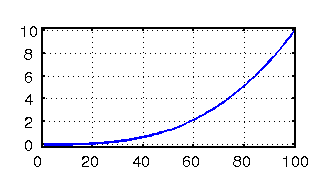}}\hspace{1pt}
        {\includegraphics[width=0.46\linewidth]{experiment_rotation/rotate_x.png}}\hspace{1pt}
        {\includegraphics[width=0.46\linewidth]{experiment_rotation/rotate_x.png}}\hspace{1pt}
        {\includegraphics[width=0.46\linewidth]{experiment_rotation/rotate_x.png}}\hspace{1pt}
    \caption{Snapshots of test video clips. Our video clips are captured by a hand-held iPhone6 with normal setting of recording. Those tagged by $*$ are provided by Ha~{\it et al.}~\cite{ha16}, and those tagged by $\dagger$ are provided by Yu~{\it et al.}~\cite{yu14}.
    }
    \label{fig:assumption}
\end{figure}
}

{\bf Computation of $\mathbf{H}_s$.} After the alignment of vertical axis, the image may suffer from geometric distortions. As suggested by Loop~{\it et al.}~\cite{loop99}, we reduce the geometric distortion by applying a shearing matrix, which is defined as
\[ {\bf H}_s = 
    \begin{bmatrix}
        s_a & s_b & 0 \\
        0 & 1 & 0 \\
        0 & 0 & 1
    \end{bmatrix}.
\]
Since the shearing transformation only relates with the $x$-coordinate of a point, it will not affect the rectification accuracy of an image. 
We denote $w$ and $h$ as the width and height of the images, respectively. Then on the slave image, $\mathbf{a} = [\frac{w-1}{2}\ 0\ 1]^{\rm T}$, $\mathbf{b} = [w-1\ \frac{h-1}{2}\ 1]^{\rm T}$, $\mathbf{c} = [\frac{w-1}{2}\ h-1\ 1]^{\rm T}$, $\mathbf{d} = [0\ \frac{h-1}{2}\ 1]^{\rm T}$ are the midpoints on its four edges, respectively.  
Let $\widehat{\mathbf{a}}$, $\widehat{\mathbf{b}}$, $\widehat{\mathbf{c}}$ and $\widehat{\mathbf{d}}$ be the mapped points after applying the transformation $\mathbf{H}_y$, and we introduce
\begin{align*}
\mathbf{u} =  \widehat{\mathbf{b}} - \widehat{\mathbf{d}} &= [u_x\ u_y\  0],\\
\mathbf{v} =  \widehat{\mathbf{a}} - \widehat{\mathbf{c}} &= [v_x\ v_y\ 0],
\end{align*}
to ease the presentation. 
Then parameters $s_a$ and $s_b$ are estimated by preserving the perpendicularity and aspect ratio of lines ${\mathbf{bd}}$ and ${\mathbf{ac}}$. Particularly, we solve for $\mathbf{H}_s$ with:
\begin{align}
(\mathbf{H}_s\mathbf{u})^{\rm T}\mathbf{H}_s\mathbf{v} &= 0, \\
\frac{(\mathbf{H}_s\mathbf{u})^{\rm T}(\mathbf{H}_s\mathbf{u})}{(\mathbf{H}_s\mathbf{v})^{\rm T}(\mathbf{H}_s\mathbf{v})} &= \frac{h^2}{w^2}.
\end{align}
According to \cite{loop99}, the solution is given by
\begin{align}
\label{eq:sa}
s_a &= \frac{h^2 u_y^2+w^2 v_y^2}{hw(u_y v_x - u_x v_y)}, \\
\label{eq:sb}
s_b &= \frac{h^2u_x u_y+w^2 v_x v_y}{hw(u_x v_y - u_y v_x)}.
\end{align}

{\bf Computation of $\mathbf{H}_k$.} Finally, a shifting matrix $\mathbf{H}_k$ is introduced to facilitate the follow-up stereo matching, which shifts the slave image horizontally to make the maximum disparity being $0$. It has the following form,
\[ {\bf H}_k = 
    \begin{bmatrix}
        1 & 0 & k \\
        0 & 1 & 0 \\
        0 & 0 & 1
    \end{bmatrix}.
\]
Let $\widetilde{x}_i$ be the $x$-coordinate of the mapped points of $\mathbf{p}_i'$ with the transformation $\mathbf{H}_s\mathbf{H}_y$. Then $k$ is simply computed by
\begin{equation}
    k = \max_{1\leq i \leq N}\{ \widetilde{x}_i - x_i\}. 
    \label{eq:k}
\end{equation}

\begin{algorithm}[t]
\caption{DSR (Direct Self-rectification)}\label{alg:sRec}
\begin{algorithmic}[1]
\STATE {\bf{Input:}} Uncalibrated stereo images
\STATE Detect and match keypoints from the input stereo images: $\{\mathbf{p}_i, \mathbf{p}_i'\}_{i=1}^{N}$ 
\STATE Initialize the maximum percentage of inliers to zero: $p_{max} = 0$
\FOR {$t = 1$ to $T$}
\STATE Randomly select $M$ pairs of matched points.
\STATE Compute a temporary alignment matrix $\mathbf{\widehat{H}}_y$ by minimizing~(\ref{eq:regress_H_y}) on the selected pairs of points.
\STATE Count the percentage of inliers $p$ using~(\ref{eq:rms}).
\IF {$p > p_{max}$}
\STATE Set $p_{max} = p$ and $\mathbf{H}_y = \mathbf{\widehat{H}}_y$
\ENDIF
\ENDFOR
\STATE Compute the elements in the shearing matrix $\mathbf{H}_s$ using~(\ref{eq:sa}) and~(\ref{eq:sb})
\STATE Compute the elements in the shifting matrix $\mathbf{H}_k$ using~(\ref{eq:k})
\STATE Compute $\mathbf{H} = \mathbf{H}_k \mathbf{H}_s \mathbf{H}_y$
\STATE {\bf{Output:}} Transformation Matrix $\mathbf{H}$
\end{algorithmic}
\end{algorithm}
\vspace{-3pt}
\subsection{The Small-drift: How small is small?}
\label{subsec:small-drift}
To verify the validity of this assumption experimentally presented in Section~\ref{subsec:lds}, we generate synthetic stereo image pairs with $12$mm baseline at a fixed scenario (to reduce the variations in keypoint detection and matching) by varying one of the five variables $[\theta_x, \theta_y, \theta_z, t_y, t_z]$ at a time. We run DSR to calculate the proportion of aligned points by setting $\epsilon=1$, $\epsilon=2$, and $\epsilon=3$. The metric will befigure
described in Section~\ref{subsec:result_selfrec}. The corresponding accuracy curves are shown in solid lines in Fig.~\ref{fig:assumption}, where the left three sub-figures plot the accuracy with varying rotation, and the right two sub-figures plot the accuracy with varying translation. Since there exist alignment errors in corresponding points themselves, we plot as a reference the accuracy curves by running calibrated rectification with ground-truth camera parameters, shown in dotted lines. From these figures, we could observe: 1) the alignment accuracy does not drop much with increased rotation angles; and 2) the alignment accuracy is more sensitive to the translation in $y-$axis, compared to the translation in $z-$axis. These well correspond to our theoretical analysis in Section~\ref{subsec:lds}. 
Based on these experiments, we empirically regard $t_y \leq 1\text{mm}$ and $t_z \leq 2\text{mm}$ as small-drift in our following experiments.

\begin{figure}[!t]
        \centering
        {\includegraphics[width=\linewidth]{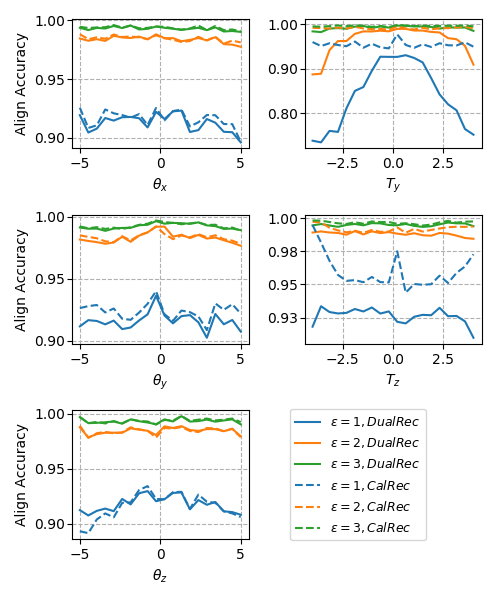}}\hspace{1pt}
    \vspace{-15pt}
    \caption{Align accuracy when varying the three rotation parameters $\theta_x, \theta_y, \theta_z$ (degree) and the two affecting translation parameters $t_y, t_z$ (mm), respectively. 
    }
    \label{fig:assumption}
\end{figure}

\section{Experiments}
\label{sec:results}

In this section, we introduce our evaluation dataset and experiment settings in Section \ref{subsec:setting}. The proposed method DSR is then evaluated in Section \ref{subsec:result_selfrec}. We further employ DSR as a pre-processing stage to other applications and demonstrate its effectiveness, shown in Section~\ref{subsec:result_bokeh}.

\subsection{Experiment Settings} 
\label{subsec:setting}
\begin{figure}[!t]
        \centering
        \subfloat[Master Image]{\includegraphics[width=0.23\linewidth]{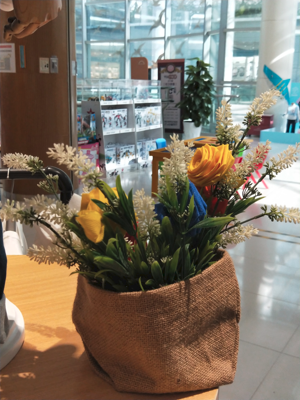}}\hspace{1pt}
        \subfloat[Slave Image]{\includegraphics[width=0.23\linewidth]{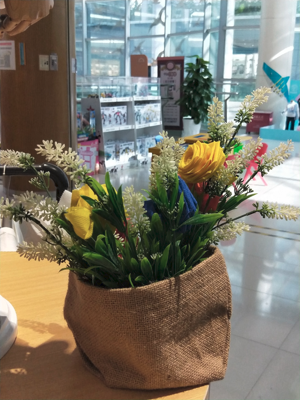}}\hspace{1pt}
        \subfloat[Master Image]{\includegraphics[width=0.23\linewidth]{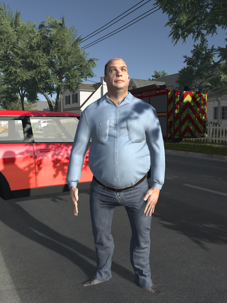}}\hspace{1pt}
        \subfloat[Slave Image]{\includegraphics[width=0.23\linewidth]{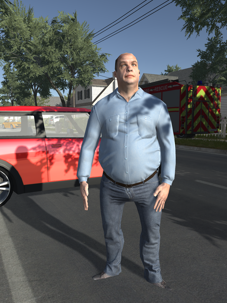}}\hspace{1pt}
    \caption{Sample images captured by dual-lens cameras at two slightly different viewpoints. A pair of synthetic images are shown in the left, and a pair of realistic images are shown in the right.}
    \label{fig:dataset}
\end{figure}

We use two datasets to evaluate our method, i)~the \emph{synthetic} dataset with simulated dual-lens camera settings and ii)~the \emph{realistic} dataset collected at various real-world scenarios. Firstly, to verify its stability across different module of mobile phones, we generated 1000 stereo pairs using Unity software~\cite{unity}, under similar settings with a real dual-lens camera on smart phones. The baseline is set to be $12$mm. To simulate the fabrication randomness, we vary the rotation angles and translations within a reasonable range. In particular, we set $\theta_x \in [-3^\circ, +3^\circ]$, $\theta_y \in [-3^\circ, +3^\circ]$, $\theta_z \in [-3^\circ, +3^\circ]$, $t_x \in [-1\text{mm}, +1\text{mm}]$,  $t_y \in [-1\text{mm}, +1\text{mm}]$,  and $t_z \in [-2\text{mm}, +2\text{mm}]$. Secondly, to validate the effectiveness on realistic image sets, we use a dual-lens smartphone with two rear-facing dual-lens cameras to collect another 1000 pairs of images at various scenarios. Some image samples are shown in Fig. \ref{fig:dataset}.

In this paper, we use BRIEF~\cite{calonder2010brief} feature for detection and matching, though one can also employ other methods such as SIFT~\cite{gkg509} or SURF~\cite{Bay2006}. We set $M = 20$, $T = 100$, $\epsilon = 1,\ 2,\ 3$, when running the proposed DSR.
\begin{figure*}[htbp]
    \centering
    \vspace{-10pt}
    {\includegraphics[width=0.19\linewidth]{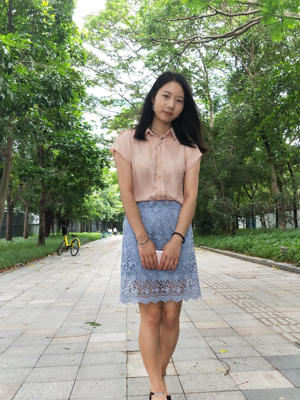}}\hspace{1pt}
    {\includegraphics[width=0.19\linewidth]{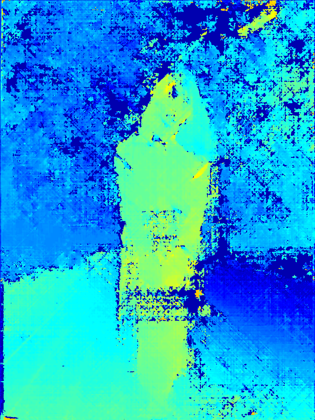}}\hspace{1pt}
    {\includegraphics[width=0.19\linewidth]{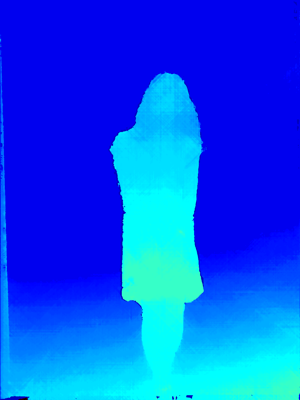}}\hspace{1pt}
    {\includegraphics[width=0.19\linewidth]{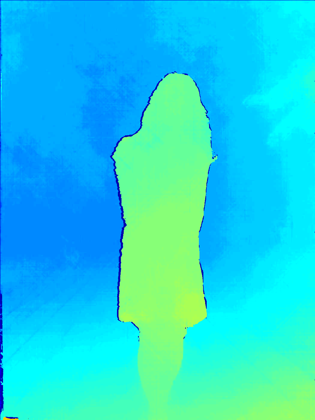}}\hspace{1pt}
    {\includegraphics[width=0.19\linewidth]{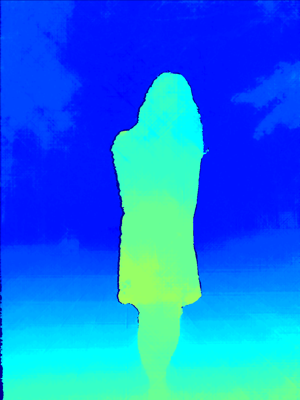}}\hspace{1pt}
    \vspace{2pt}\\
    \subfloat[Master Image]
    {\includegraphics[width=0.19\linewidth]{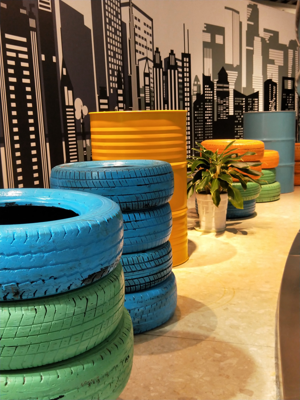}}\hspace{1pt}
    \subfloat[\texttt{CalRec+SGM}]
    {\includegraphics[width=0.19\linewidth]{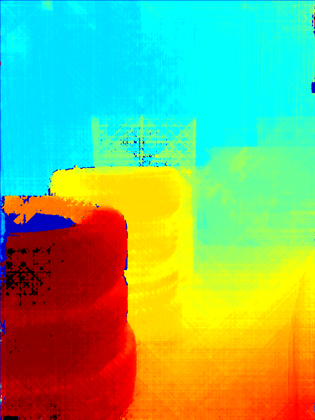}}\hspace{1pt}
    \subfloat[\texttt{Loop}~\cite{loop99}\texttt{+SGM}]
    {\includegraphics[width=0.19\linewidth]{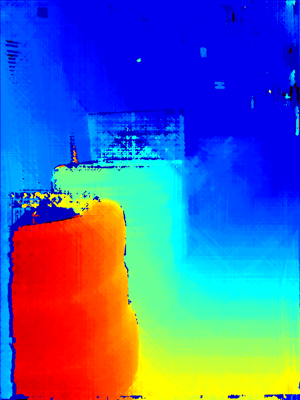}}\hspace{1pt}
    \subfloat[\texttt{CalRec+DSR+SGM}]
    {\includegraphics[width=0.19\linewidth]{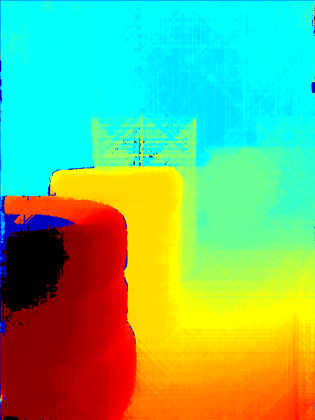}}\hspace{1pt}
    \subfloat[\texttt{DSR+SGM}]
    {\includegraphics[width=0.19\linewidth]{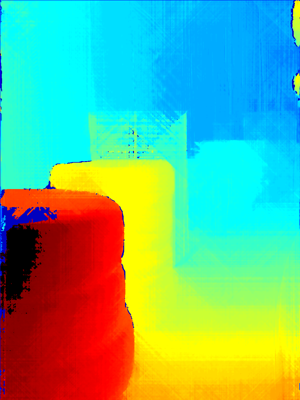}}\hspace{1pt}
    \caption{Comparisons of stereo matching results with different rectification algorithm to pre-process stereo images from the realistic dataset. Best viewed in color.
    }
    \label{fig:stereo_disp_compare}
\end{figure*}

\subsection{Performance of DSR} 
\label{subsec:result_selfrec}
{\bf Evaluation Metrics.}
We evaluate the rectification algorithms by two metrics. The first metric is the vertical alignment accuracy, which is defined as the proportion of well aligned points (short for PAP):
\begin{equation}
\text{PAP} =  \frac{\sum_{i=1}^{N}\mathbf{1}(|y_{i} - \widetilde{y}_{i}| < \epsilon)}{N}
\label{eq:rms}
\end{equation}
where $y_{i}$ and $\widetilde{y}_{i}$ are the $y$-coordinate of the $i^{th}$ pair of corresponding points in two images , $N$ is the total number of corresponding pairs, $\mathbf{1}(\mathord{\cdot})$ is the indicator function and $\epsilon$ has the same value as the one in DSR algorithm. $\epsilon = 1,\ 2,\ 3$ is evaluated in our experiments. 
Conceptually similar to the commonly used {\it reprojection error} in camera calibration, the PAP  metric also evaluates the alignment accuracy. However, as outliers inevitably exist during evaluation, though a very small portion, we use the proportion of aligned points rather than the distance of re-projected points to reduce the influence introduced by the inevitable outliers.
The second metric is defined to measure the geometric distortion on an image by a normalized vertex distance. Let $\mathbf{v}_1= [0\ 0\ 1]$, $\mathbf{v}_2= [w-1\ 0\ 1]$, $\mathbf{v}_3= [0\ h-1\ 1]$ and $\mathbf{v}_4= [w-1\ h-1\ 1]$ be the four vertices of an image, $w$ and $h$ be the width and height of the image, respectively. Then
\begin{equation}
    \text{NVD} =  \frac{\sum_{i=1}^{4}d_i}{\sqrt{w^2+h^2}},
\end{equation}
where $d_i$ is the Euclidean distance between $\mathbf{v}_i$ and its transformed point.

{\bf Comparison and Evaluation.}
We compare with two other methods. Given a set of offline calibrated camera parameters, the first rectification method we compared is the widely used OpenCV~\cite{opencv_library} implementation of calibrated rectification, which will be referred to as \texttt{ CalRec} in the rest of this paper. The second one is a classic rectification algorithm proposed by Loop~{\it et al.}~\cite{loop99}. We also tried to evaluate the 
newer algorithms proposed by Fusiello~{\it et al.}~\cite{fusiello2011quasi}. However, when running over the whole dataset, we observe that the implementation provided by the authors is unstable due to the use of Levenberg-Marquardt with all the unknown variables set to zero at the beginning. Thus we do not report the results here.
For a fair comparison, both methods ({\it i.e.,} Loop's~\cite{loop99} and ours) share the same strategy and the same parameters for corresponding points extraction. In addition, the same RANSAC is applied to make both methods robust  to  possible  outliers. 
The evaluation results for the synthetic dataset and realistic dataset are shown in Table~\ref{tab:rec_obj}. As observed from the table, our method greatly reduces the vertical alignment error as well as geometric distortion  compared to the method of Loop~\cite{loop99}. Idealy, \texttt{CalRec} for both synthetic and realistic dataset should have the best PAP, given that the camera parameters are calibrated. However, it is common that a smartphone may have different camera parameters with its initially fabricated ones, due to change of focal length, movement of camera modules, {\it etc.}, which degrades the usage of calibrated data.

\begin{table}[htbp]
\small
\centering
\begin{tabular}{c|c|c|c|c|c}
\cline{1-6} 
\multirow{2}{*}{Methods} & \multicolumn{3}{c|}{PAP} & \multicolumn{2}{c}{NVD} \\ \cline{2-6} 
& $\epsilon = 1$ & $\epsilon = 2$ & $\epsilon = 3$ & master & slave \\ \cline{1-6} 
&\multicolumn{5}{c}{\phantom{bl} Evaluation on synthetic dataset\phantom{bl}}\\ \cline{1-6}
CalRec & \bf{0.8242} & {\bf 0.9404} & {\bf 0.9628} & 0.2054 & 0.2273\\ \cline{1-6}
Loop~\cite{loop99} & 0.6006 & 0.8539 & 0.9404 & 0.1423 & 0.1658 \\ \cline{1-6} 
DSR (Ours) & 0.8087 &  0.9370 & {\bf 0.9628} & {\bf 0.0000} & {\bf 0.1652}\\ \cline{1-6} 
& \multicolumn{5}{c}{\phantom{bl} Evaluation on realistic dataset\phantom{bl}}\\ \cline{1-6}
\cline{1-6} 
CalRec & 0.2811 & 0.5125 & 0.7317 & 0.0741 & 0.1274 \\ \cline{1-6}
Loop~\cite{loop99} & 0.7624 & 0.9404 & 0.9634 &0.0276& 0.0337\\ \cline{1-6} 
DSR (Ours) & \bf{0.8324} & \bf{0.9501} & \bf{0.9732} &\bf{0.0000} & \bf{0.0063}\\ \cline{1-6} 
\end{tabular}
\vspace{5pt}
\caption{Quantitative evaluation of rectification algorithms by Percentage of Aligned Points (PAP), and Normalized Vertex Distance (NVD). Higher PAP indicates better accuracy, lower NVD indicates better accuracy.}
\label{tab:rec_obj}
\end{table}

On a desktop equipped with Intel I5 CPU, the running time of two methods, \texttt{Loop}~\cite{loop99} and our \texttt{DSR}, is reported in Table~\ref{tab:time} with the input image resolution being  720$\times$960. The homography estimation in DSR is extremely fast with feature matching being the bottleneck. We believe it can be further accelerated by parallel processing languages or optimized for customized platforms.
\begin{table}[htbp]
\small
\centering
\begin{tabular}{c|c|c}
\cline{1-3} 
Methods & Feature Matching & Homography Estimation \\ \cline{1-3} 
\texttt{Loop}~\cite{loop99} & 428.10 ms & 59.51 ms \\ \cline{1-3} 
\texttt{DSR} (Ours) & 428.10 ms & 0.63 ms\\ \cline{1-3} 
\end{tabular}
\vspace{0.5pt}
\caption{Running time of \texttt{Loop}~\cite{loop99} and our \texttt{DSR}, measured in milli-second (ms).}
\label{tab:time}
\end{table}


\subsection{Applications on Depth-of-field Rendering}
\label{subsec:result_bokeh}

\begin{figure}[htbp]
        \centering
        \vspace{-10px}
        {\includegraphics[width=0.48\linewidth]{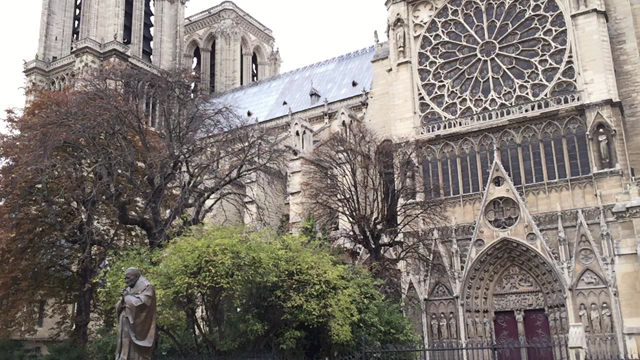}}\hspace{1pt}
        {\includegraphics[width=0.48\linewidth]{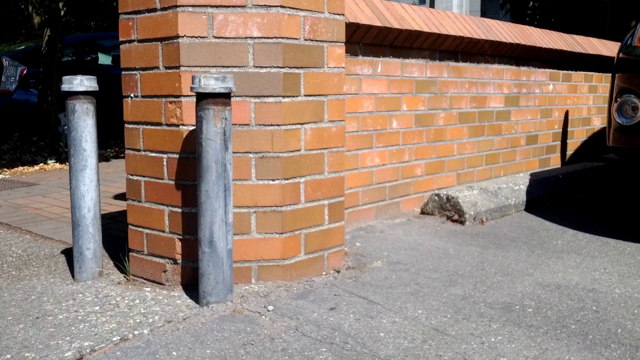}}\hspace{1pt}
        {\includegraphics[width=0.48\linewidth]{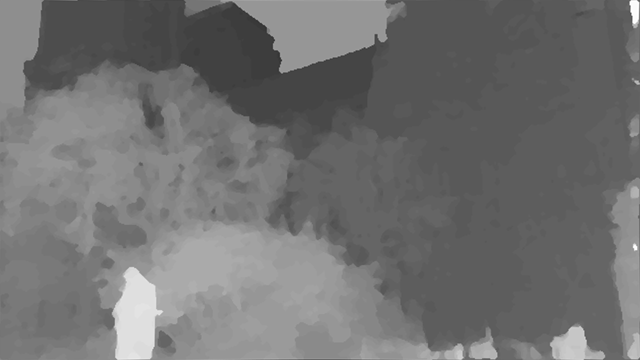}}\hspace{1pt}
        {\includegraphics[width=0.48\linewidth]{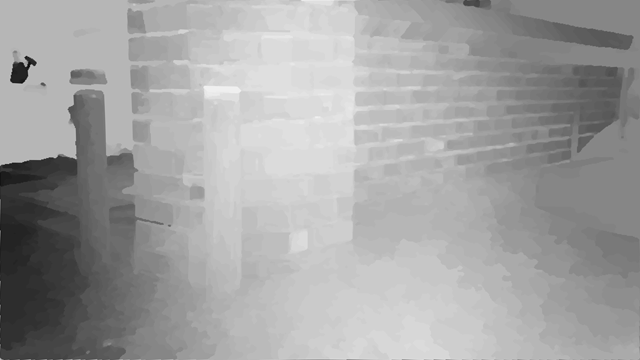}}\hspace{1pt}
        {\includegraphics[width=0.48\linewidth]{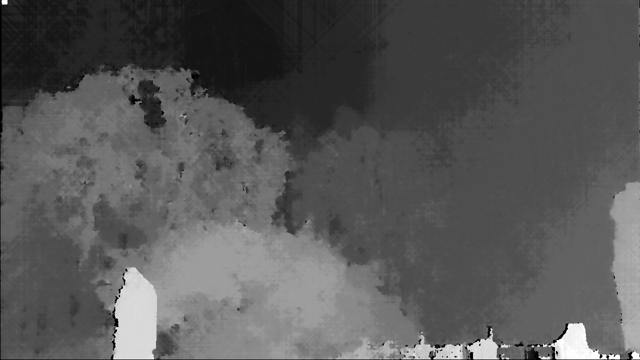}}\hspace{1pt}
        {\includegraphics[width=0.48\linewidth]{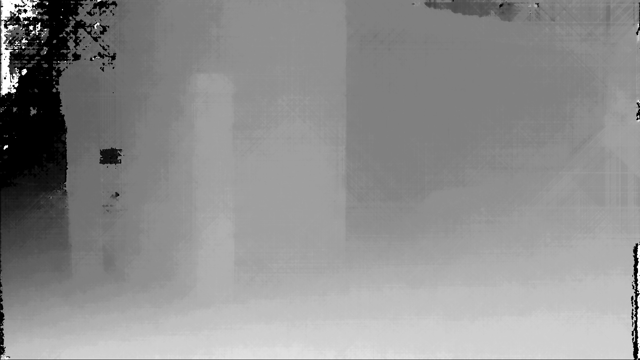}}\hspace{1pt}
    \caption{Comparisons of the disparity maps. The ones shown in the second row are generated by \texttt{Ha}~\cite{ha16}, and the two disparity maps  shown in the third row are generated by \texttt{DSR+SGM} with no additional depth refinement.}
    \label{fig:disp_compare_video}
\end{figure}

{\bf Applied to Stereo Matching.} As rectification is usually served as a pre-processing to stereo matching algorithms, it is worthy to know whether the rectified stereo images can bring satisfactory stereo matching results. To do so, we choose the commonly adopted semi-global matching (SGM)~\cite{hirschmuller08} from a list of stereo matching algorithms~\cite{scharstein02, pang17, zagoruyko15, hirschmuller08}, and compare the above mentioned three rectification algorithms. For ease of convenience, we denote the combined algorithm flow as \texttt{CalRec+SGM}, \texttt{Loop}~\cite{loop99}\texttt{+SGM}, and \texttt{DSR+SGM}. The estimated disparity maps are shown in Fig.~\ref{fig:stereo_disp_compare}. Notice that to analyze qualitatively the quality of rectification algorithms, we did not implement any refinement techniques on the results of \texttt{SGM}. As can be observed, \texttt{DSR+SGM} has less mis-calculated pixels compared to \texttt{Loop}~\cite{loop99}\texttt{+SGM} and \texttt{CalRec+SGM}. Notice that as there are geometic distortions after image warping of \texttt{CalRec} and \texttt{Loop}~\cite{loop99}, the pixels at image boundaries are erroneously estimated. To test the ability to compensate the calibration errors after fabrication (also mentioned in Section~\ref{subsec:result_selfrec}), we also show the disparity maps of \texttt{CalRec + DSR + SGM} in Fig.~\ref{fig:stereo_disp_compare}. Clearly, the noises and inaccurate regions on the disparity map greatly reduces.

\begin{figure}[htbp]
        \centering
        \vspace{-10px}
        {\includegraphics[width=0.45\linewidth]{figures/disp/1755_selfrec_out_image0.png}}\hspace{1pt}
        {\includegraphics[width=0.45\linewidth]{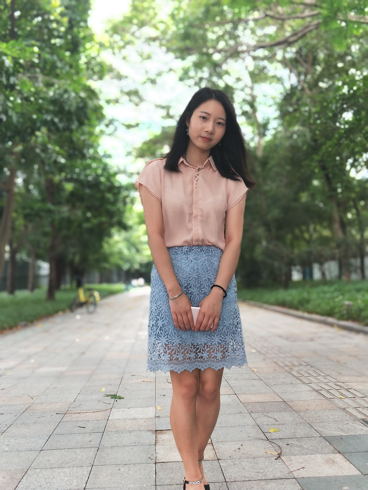}}\hspace{1pt}
        \vspace{5pt}
        {\includegraphics[width=0.45\linewidth]{figures/disp/3853_selfrec_out_image0.png}}\hspace{1pt}
        {\includegraphics[width=0.45\linewidth]{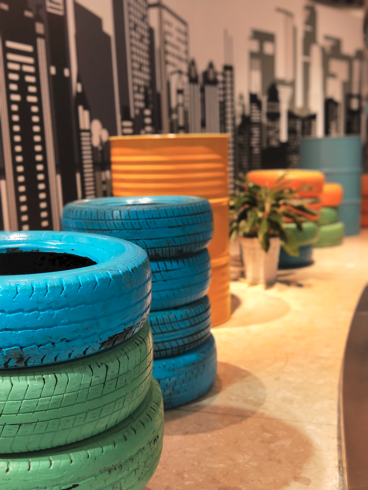}}\hspace{1pt}
    \caption{Depth-of-field rendering results based on the rectified images processed by the proposed \texttt{DSR}. Best for enlarged views.}
    \label{fig:bokeh}
\end{figure}
In Figure~\ref{fig:disp_compare_video}, we compare \texttt{DSR+SGM} with the methods proposed by \texttt{Ha}~\cite{ha16} for uncalibrated small motion clip on their collected dataset. To run \texttt{DSR}, we select only two frames with relatively higher alignment accuracy. Compared with Ha~\cite{ha16}, \texttt{DSR+SGM} generates more stable and accurate disparity maps. The quantitative assessment of the estimated disparities are not performed, since the ground-truth disparities are not available. 

{\bf Applied to Depth-of-field Rendering.} Fig. \ref{fig:bokeh} shows the depth-of-field rendering~\cite{kraus07} results applied to the master image where the kernel size is estimated based on the computed depth. WLS filtering~\cite{min14} is utilized to refine the estimated disparity maps with edge preserving properties.

\section{Discussion and Conclusion}
\label{sec:conclude}
We present a self-rectification approach called DSR for uncalibrated dual-lens smartphone cameras. The proposed DSR achieves superb accuracy in terms of percentage of aligned points (PAP) and zero geometric distortion for master image in terms of normalized vertex distance (NVD). The effectiveness is further validated by applying stereo matching and depth-of-field rendering on the rectified image pairs. 
As DSR is designed for dual-lens cameras with small-drift properties, the method is not suggested to rectify stereo image pairs with large translation. Fortunately, almost all types of dual-lens smartphones can benefit from the high effectiveness of the proposed algorithm. Though some dual-lens cameras have wide-and-tele  or color-and-gray cameras, as long as sufficient keypoints can be matched, DSR can be employed to rectify the stereo images. DSR may fail when insufficient correct keypoints are detected, for example a stereo image pair of an entire textureless white wall.
The proposed DSR can be applied as a pre-processing step  to stereo matching for a wide range of applications, such as depth-of-field rendering, 3D segmentation, portrait relighting. It can also be applied to generate training samples for unsupervised / semi-supervised stereo matching networks.

{
\bibliographystyle{ieee}
\bibliography{ref}

\begin{thebibliography}{10}\itemsep=-1pt

\bibitem{unity}
Unity.
\newblock \url{https://unity3d.com}.
\newblock Accessed: 2017-11-14.

\bibitem{Bay2006}
H.~Bay, T.~Tuytelaars, and L.~Van~Gool.
\newblock {\em SURF: Speeded Up Robust Features}, pages 404--417.
\newblock Springer Berlin Heidelberg, Berlin, Heidelberg, 2006.

\bibitem{opencv_library}
G.~Bradski.
\newblock {The OpenCV Library}.
\newblock {\em Dr. Dobb's Journal of Software Tools}, 2000.

\bibitem{calonder2010brief}
M.~Calonder, V.~Lepetit, C.~Strecha, and P.~Fua.
\newblock Brief: Binary robust independent elementary features.
\newblock {\em Computer Vision--ECCV 2010}, pages 778--792, 2010.

\bibitem{fusiello2011quasi}
A.~Fusiello and L.~Irsara.
\newblock Quasi-euclidean epipolar rectification of uncalibrated images.
\newblock {\em Machine Vision and Applications}, 22(4):663--670, 2011.

\bibitem{fusiello00}
A.~Fusiello, E.~Trucco, and A.~Verri.
\newblock A compact algorithm for rectification of stereo pairs.
\newblock {\em Machine Vision and Applications}, 12(1):16--22, 2000.

\bibitem{gluckman01}
J.~Gluckman and S.~K. Nayar.
\newblock Rectifying transformations that minimize resampling effects.
\newblock In {\em Computer Vision and Pattern Recognition, 2001. CVPR 2001.
  Proceedings of the 2001 IEEE Computer Society Conference on}, volume~1, pages
  I--I. IEEE, 2001.

\bibitem{ha16}
H.~Ha, S.~Im, J.~Park, H.-G. Jeon, and I.~So~Kweon.
\newblock High-quality depth from uncalibrated small motion clip.
\newblock In {\em Proceedings of the IEEE Conference on Computer Vision and
  Pattern Recognition}, pages 5413--5421, 2016.

\bibitem{hartley2003multiple}
R.~Hartley and A.~Zisserman.
\newblock {\em Multiple view geometry in computer vision}.
\newblock Cambridge university press, 2003.

\bibitem{hirschmuller08}
H.~Hirschmuller.
\newblock Stereo processing by semiglobal matching and mutual information.
\newblock {\em IEEE Transactions on pattern analysis and machine intelligence},
  30(2):328--341, 2008.

\bibitem{isgro99}
F.~Isgro and E.~Trucco.
\newblock Projective rectification without epipolar geometry.
\newblock In {\em Computer Vision and Pattern Recognition, 1999. IEEE Computer
  Society Conference on.}, volume~1, pages 94--99. IEEE, 1999.

\bibitem{kraus07}
M.~Kraus and M.~Strengert.
\newblock Depth-of-field rendering by pyramidal image processing.
\newblock In {\em Computer Graphics Forum}, volume~26, pages 645--654, 2007.

\bibitem{loop99}
C.~Loop and Z.~Zhang.
\newblock Computing rectifying homographies for stereo vision.
\newblock In {\em Computer Vision and Pattern Recognition, 1999. IEEE Computer
  Society Conference on.}, volume~1, pages 125--131. IEEE, 1999.

\bibitem{min14}
D.~Min, S.~Choi, J.~Lu, B.~Ham, K.~Sohn, and M.~N. Do.
\newblock Fast global image smoothing based on weighted least squares.
\newblock {\em IEEE Transactions on Image Processing}, 23(12):5638--5653, 2014.

\bibitem{gkg509}
P.~C. Ng and S.~Henikoff.
\newblock Sift: predicting amino acid changes that affect protein function.
\newblock {\em Nucleic Acids Research}, 31(13):3812--3814, 2003.

\bibitem{pang17}
J.~Pang, W.~Sun, J.~S. Ren, C.~Yang, and Q.~Yan.
\newblock Cascade residual learning: A two-stage convolutional neural network
  for stereo matching.
\newblock In {\em ICCV Workshop on Geometry Meets Deep Learning}, Oct 2017.

\bibitem{scharstein02}
D.~Scharstein and R.~Szeliski.
\newblock A taxonomy and evaluation of dense two-frame stereo correspondence
  algorithms.
\newblock {\em International journal of computer vision}, 47(1-3):7--42, 2002.

\bibitem{zagoruyko15}
S.~Zagoruyko and N.~Komodakis.
\newblock Learning to compare image patches via convolutional neural networks.
\newblock In {\em Proceedings of the IEEE Conference on Computer Vision and
  Pattern Recognition}, pages 4353--4361, 2015.

\bibitem{zhang98}
Z.~Zhang.
\newblock Determining the epipolar geometry and its uncertainty: A review.
\newblock {\em International journal of computer vision}, 27(2):161--195, 1998.

\bibitem{zilly10}
F.~Zilly, M.~M{\"u}ller, P.~Eisert, and P.~Kauff.
\newblock Joint estimation of epipolar geometry and rectification parameters
  using point correspondences for stereoscopic tv sequences.
\newblock In {\em Proceedings of 3DPVT}, 2010.

\end{thebibliography}
}

\end{document}